\documentclass[10pt, a4paper]{article}
\usepackage{lrec2022} 
\usepackage{multibib}
\newcites{languageresource}{Language Resources}
\usepackage{graphicx}
\usepackage{tabularx}
\usepackage{soul}
\usepackage{graphicx,multirow}
\usepackage{amsmath}

\usepackage{titlesec}
\titleformat{\section}{\normalfont\large\bfseries\center}{\thesection.}{1em}{}
\titleformat{\subsection}{\normalfont\SmallTitleFont\bfseries\raggedright}{\thesubsection.}{1em}{}
\titleformat{\subsubsection}{\normalfont\normalsize\bfseries\raggedright}{\thesubsubsection.}{1em}{}
\renewcommand\thesection{\arabic{section}}
\renewcommand\thesubsection{\thesection.\arabic{subsection}}
\renewcommand\thesubsubsection{\thesubsection.\arabic{subsubsection}}

\usepackage{epstopdf}
\usepackage[utf8]{inputenc}

\usepackage{hyperref}
\usepackage{xstring}

\usepackage{color}

\title{An Evaluation of Sindhi Word Embedding in Semantic Analogies and Downstream Tasks}

\name{Wazir Ali$^{1}$$,
 \textbf{Saifullah Tumrani}$$^{2}$$, 
 \textbf{Jay Kumar}$$^{3}$$, 
 \textbf{Tariq Rahim Soomro}$$^{1}$}

\address{
    $^{1}$ College of Computer Science and Information Systems, Institute of Business Management, 75190 Karachi, Pakistan\\
    $^{2}$ Department of Bioinformatics, Heidelberg University, Germany\\
    $^{3}$ Faculty of Computer Science, Dalhousie University, Halifax, NS, Canada\\
    \texttt{aliwazirjam@gmail.com}
}

\abstract{
In this paper, we propose a new word embedding based corpus consisting of more than 61 million words crawled from multiple web resources.  We design a preprocessing pipeline for the filtration of unwanted text from crawled data. Afterwards, the cleaned vocabulary is fed to state-of-the-art continuous-bag-of-words, skip-gram, and GloVe word embedding algorithms. For the evaluation of pretrained embeddings, we use popular intrinsic and extrinsic evaluation approaches. The evaluation results reveal that continuous-bag-of-words and skip-gram perform better than GloVe and existing Sindhi fastText word embedding on both intrinsic and extrinsic evaluation approaches.
 \newline \Keywords{Language Resources, Corpus Acquisition, Sindhi Language, Neural Networks, Word Embeddings } }

\begin{document}
\maketitle
\section{Introduction}
\label{Intro}
 Sindhi is a rich morphological, multi-script, and multi-dialectal language. It belongs to the Indo-Aryan language family \cite{cole2006sindhi}, with significant cultural and historical background. Presently, it is recognized as is an official language \cite{motlani2016developing} in Sindh province of Pakistan, also being taught as a compulsory subject in schools and colleges.  Sindhi is also recognized as one of the national languages in India \cite{ali2015towards}. It is spoken by nearly 75 million people \cite{motlani2016developing}. Persian-Arabic is the standard script of Sindhi, which was officially accepted in 1852 by the British government\footnote{\url{https://www.britannica.com/topic/Sindhi-language}}. However, the Sindhi-Devanagari is also a popular writing system in India being written in left to right direction like Hindi language. Sindhi stands among the low-resource languages due to the scarcity of core language resources (LRs) of the unlabelled corpus, which can be utilized for training word embeddings or state-of-the art language models. 


Language resources are fundamental elements for the development of high-quality natural language processing (NLP) systems. The development of such resources has received great research interest for the digitization of human languages~\cite{ali2020siner}. Such LRs include written or spoken corpora, lexicon, and annotated corpora. Many world languages are rich in such LRs, including English \cite{spacy,nltk,manning2014stanford}, Chinese \cite{che2010ltp} and other languages \cite{popel2010tectomt,padro2010freeling}. Sindhi is still at its developing phase for its basic LRs~\cite{ali2021sipos}. Only a few resources have been introduced for Sindhi Persian-Arabic including raw corpus~\cite{rahman2010towards,bhatti2014word,dootio2019development,motlani2016developing} , labelled corpus~\cite{ali2017sentiment,dootio2019syntactic,ali2020siner,ali2021sipos,ali2021creating}. Unfortunately, the existing raw corpora are not sufficient to train  word embeddings and developing language-independent NLP applications for statistical Sindhi language processing such as semantic, semantic analysis and automatic development of WordNet. 

More recently, neural network based models~\cite{otter2020survey} yield state-of-the-art performance in NLP with the word embeddings~\cite{pennington2014glove,mikolov2018advances,grave2018learning}. One of the advantages of such techniques is they use unsupervised approaches for learning representations and do not require annotated corpus, which is rare for low-resourced Sindhi language. 

In this paper, we address the problem of corpus scarcity by crawling a large corpus of more than 61 million words from multiple web resources using the web-scrappy. Afterwards, the corpus is utilized to train Sindhi word embeddings using state-of-the-art GloVe \cite{pennington2014glove}, SG and CBoW \cite{mikolov2013efficient,mikolov2013distributed,mikolov2018advances} algorithms. The popular intrinsic~\cite{schnabel2015evaluation} and extrinsic~\cite{nayak2016evaluating} evaluation methods are employed for the performance evaluation of the proposed and existing fastText~\cite{grave2018learning} Sindhi word embeddings. The synopsis of our novel contributions is listed as follows:
 
\begin{itemize}
	\item We crawl a large corpus of more than 61 million words obtained from multiple web resources and reveal a list of Sindhi stop-words.
	\item We generate word embeddings using CBoW, SG, and GloVe. Evaluate them using the popular intrinsic and extrinsic evaluation approaches.
	\item  We evaluate Sindhi fastText word representations and compare them with our proposed Sindhi word embeddings. 
\end{itemize}
 
\section{Related work}
\label{relatedwork}
Abundant LRs are available for resource enriched languages which are integrated in the software tools such as Natural Language Toolkit (NLTK) for English~\cite{nltk} that provides more than  50 lexical resources and corpora  for classification, tokenization, stemming, tagging, parsing, and semantic reasoning with an easy-to-use interfaces. In contrast, Sindhi language is at early stage \cite{dootio2019development} for the development of such resources and software tools. 

The corpus construction for NLP mainly involves important steps of acquisition, preprocessing, and tokenization. Little work exists on the corpus construction for Sindhi.  Initially, \cite{rahman2010towards,khoso2019build} discussed the morphological structure and challenges concerned with the corpus development along with orthographic and morphological features in Persian-Arabic script. \cite{bhatti2014word} utilized the raw corpus for Sindhi word segmentation using a dictionary-based approach. \cite{motlani2016developing} crawled raw corpus for the annotation purpose. \cite{dootio2017automatic} proposed a basic preprocessing model for lemmatization and stemming for Sindhi Persian Arabic. \cite{nathani2019design} also proposed lemmatizer for Sindhi-Devanagari. \cite{shah2018designing} build a corpus of more than 1 million words by crawling news, books, magazines, and blogs for the ongoing annotation project for Sindhi using  XML tagging approach. The corpus lacks open-source availability, and the statistics do not show the text analysis. \cite{dootio2019development} collected and analyzed text corpus from multiple web resources. The analysis is performed by using N-grams for term frequency-inverse document frequency and document term matrix. We present the gist of existing and proposed resources in Table \ref{tab:related_work} on the corpus development and word embeddings, respectively.

Word embeddings such as fastText~\cite{bojanowski2017enriching,mikolov2018advances}, Word2Vec~\cite{mikolov2013distributed,mikolov2013efficient}, and GloVe~\cite{pennington2014glove} are popular and key methods to solving many NLP problems. 

The recent use cases of word embeddings are not only limited to boost statistical NLP applications but can also be used to develop other LRs, such as automatic construction of WordNet \cite{khodak2017automated} using the unsupervised approach. The performance of word embeddings is evaluated using the intrinsic evaluation methods \cite{schnabel2015evaluation,pierrejean2018towards}  and extrinsic evaluation methods~\cite{nayak2016evaluating}. The intrinsic approach is used to measure the internal quality of word embeddings, such as by querying nearest neighboring words \cite{pierrejean2018towards} and calculating the semantic or syntactic similarity between similar word pairs. The key advantage of that method is to reduce bias and create insight to find data-driven relevance judgment. An extrinsic evaluation approach is used to evaluate the performance in downstream NLP tasks, such as POS tagging or named-entity recognition \cite{schnabel2015evaluation}. But Sindhi lacks the unlabelled corpus, which can be utilized for training word embeddings.

\begin{table}[]
\centering
    {
    $\begin{tabular}{|l|l|l|}
    \hline
    & \textbf{Paper} &	\textbf{Resource} \\
     \hline
     \multirow{4}{*}{\rotatebox[origin=c]{90}{~Corpus}} 
     & \cite{rahman2010towards} & 4.1M tokens   \\
     \cline{2-3}
		& \cite{bhatti2014word} &   1,575K tokens  \\  \cline{2-3}
		& \cite{motlani2016developing} &  Wiki-dumps (2016) \\ \cline{2-3}
		& \cite{shah2018designing} & 100K tokens\\
		\cline{2-3}
		& \cite{dootio2019development} & 31K tokens\\ \cline{2-3}
		\hline
     \multirow{5}{*}{\rotatebox[origin=c]{90}{~Embedding}} 
     & \cite{grave2018learning} (fastText) & Wiki-dumps (2016)\\ \cline{2-3}
		& This work  & 61.39M tokens\\ 
		& (SG, CBoW, GloVe) &\\ \cline{2-3}
		& This work  & 340 stop words\\ \cline{2-3}
		& Sindhi WordSim  & 347 word pairs\\
     \hline
    \end{tabular}$
    }
\caption{Comparison of the existing and proposed work on Sindhi corpus construction and word embeddings. The stop-words were filtered from the entire vocabulary for the training of GloVe model.}
\label{table:final_results}
\end{table}

\section{Corpus Acquisition}
\label{sec:methodology}
The corpus is a collection of human language text \cite{schafer2013web} built with a specific purpose. We initiate this study from scratch by crawling large corpus from multiple web resources. This section presents the employed methodology in detail for corpus acquisition, preprocessing, statistical analysis. In fact, realizing the necessity of large text corpus for Sindhi, we started this research by collecting raw corpus from multiple web resource using web-scrappy framwork\footnote{\url{https://github.com/scrapy/scrapy}} for extraction of news columns of daily Kawish\footnote{\url{http://kawish.asia/Articles1/index.htm}} and Awami Awaz\footnote{\url{http://www.awamiawaz.com/articles/294/}} Sindhi newspapers, Wikipedia dumps\footnote{\url{https://dumps.wikimedia.org/sdwiki/20180620/}}, short stories and sports news from Wichaar\footnote{\url{http://wichaar.com/news/134/}} social blog, news from Focus Word press blog\footnote{\url{https://thefocus.wordpress.com/}}, historical writings, novels, stories, books from Sindh Salamat\footnote{\url{http://sindhsalamat.com/}} literary websites, novels, history and religious books from Sindhi Adabi Board \footnote{\url{http://www.sindhiadabiboard.org/catalogue/History/Main_History.HTML}} and tweets regarding news and sports are collected from twitter\footnote{\url{https://twitter.com/dailysindhtimes}}.

\begin{table*}
	\centering
	\begin{tabular}{|l|l|l|l|l|}
		\hline
		\textbf{Source} &  \textbf{Category}  &  \textbf{Sentences} &      \textbf{Vocabulary} & \textbf{Unique words} \\
		\hline
		Kawish & News columns  &    473,225 & 13,733,379 &    109,366 \\
		\hline
		Awami awaz & News columns &    107,326 &  7,487,319 &     65,632 \\
		\hline
		Wikipedia & Miscellaneous &    844,221 &  8,229,541 &    245,621 \\
		Social Blogs & Stories, sports  &      7,018 &    254,327 &     10,615 \\
		& History, News &      3,260 &    110,718 &      7,779 \\
		\hline
		Focus word press & Short Stories &     63,251 &    968,639 &     28,341 \\
		&     Novels &     36,859 &    998,690 &     18,607 \\
		&  Safarnama &    138,119 &  2,837,595 &     53,193 \\
		\hline
		Sindh Salamat &    History &    145,845 &  3,493,020 &     61,993 \\
		&   Religion &     96,837 &  2,187,563 &     39,525 \\
		&    Columns &     85,995 &  1,877,813 &     33,127 \\
		& Miscellaneous  &    719,956 &  9,304,006 &    168,009 \\
		\hline
		Sindhi Adabi Board & History books &    478,424 &  9,757,844 &     57,854 \\
		Twitter  & News tweets &     10,752 &    159,130 &      9,794 \\
		\hline
		Total &            &  3,211,088 & 61,399,584 &     908,456 \\
		\hline
	\end{tabular}
	\caption{Statistics of crawled corpus from multiple web resources}
	\label{tab:dataset_statistics}
\end{table*}

\subsection{Preprocessing} 
\label{sec:pre-processing}
The preprocessing of text corpus obtained from multiple web resources is a challenging task especially it becomes more complicated when working on low-resourced language like Sindhi due to the lack of open-source preprocessing tools such as NLTK \cite{nltk} for English. Therefore, we design a preprocessing pipeline for the filtration of unwanted data and vocabulary of other languages such as English in order to prepare input for word embedding models. Moreover, we reveal the list of Sindhi stop-words, which is a labor-intensive task and requires human judgment as well. Hence, the most frequent and least important words are classified as stop-words with the help of a Sindhi linguistic expert.
\begin{itemize}
	\item \textbf{Input:} The first part of the preprocessing pipeline consists of the input of collected text documents after concatenation.
	\item \textbf{Replacement symbols:} The punctuation marks, such as comma, quotation, exclamation, colon, and semi-colon are replaced with white space for authentic tokenization. Without replacing these symbols with white space, the words were joined with their next or previous corresponding words. 
	
	\item \textbf{Tokenization:} The punctuation marks, \textit{except full stop and question mark}, special symbols were replaced with white space for authentic tokenization. Afterwards, the white spaces and punctuation markers \textit{full stop and question mark} are used as word boundaries for the tokenization.
	
	\item \textbf{Filtration of noisy data:} The text acquisition from web resources contains a huge amount of noisy data. After the tokenization process, the noisy data is filtered out, such as the rest of the punctuation marks, special characters, HTML tags, all types of numeric entities, math symbols, email, and web addresses. 
	
	\item \textbf{Normalization:} In this step, we normalize the text to lower-case for the filtration of English vocabulary, and duplicate words. 

	\item \textbf{Output:} We obtain the cleaned vocabulary after employing preprocessing pipeline on the crawled data. 
	\end{itemize}

\section{Corpus Statistics}
\label{sec:statistical_analysis_of_corpus}
The large corpus acquired from multiple resources is rich in vocabulary. We present the complete statistics of collected corpus (see Table \ref{tab:dataset_statistics}) with number of sentences, words and unique tokens.

\subsection{Letter occurrences}
\label{sec:letter_occurances}
The length of words is analyzed, which is essential to develop NLP systems, including learning of word embeddings to choose the minimum or maximum length of subwords for character-level representation learning \cite{mikolov2018advances}. The table Table \ref{tab:letter_ngrams} shows that bi-gram words are most frequent, mostly consist of stop-words. Secondly, 4-gram words have a higher frequency in the corpus. 

\begin{table}
	\centering
	\begin{tabular}{|l|l|l|}
		\hline
		n-grams  & Frequency  & \% in corpus \\
		\hline
		Uni-gram &       936,301 &    1.52889 \\
		Bi-gram &    19,187,314 &    31.3311 \\
		Tri-gram &   11,924,760 &     19.472 \\
		4-gram &     14,334,444 &    23.4068 \\
		5-gram &     9,459,657 &    15.4467 \\
		6-gram &     3,347,907 &     5.4668 \\
		7-gram &     1,481,810 &     2.4196 \\
		8-gram &       373,417 &     0.6097 \\
		9-gram &       163,301 &     0.2666 \\
		10-gram &      21,287 &     0.0347 \\
		11-gram &      5,892 &     0.0096 \\
		12-gram &       3,033 &     0.0049 \\
		13-gram &       1,036 &     0.0016 \\
		14-gram &        295 &     0.0004 \\
		\hline
		Total &     61,240,454 &        100 \\
		\hline
	\end{tabular}  
		\caption{Length of letter n-grams in words, along-with frequency and percentage in corpus}
	\label{tab:letter_ngrams}
\end{table}

\subsection{Stop words}
\label{sec:stop_words}
The most frequent and least important words in NLP are often classified as stop-words. The removal of such words can boost the performance of the NLP model \cite{pandey2009evaluating}, such as sentiment analysis and text classification. But the construction of such a word list is time-consuming and requires user decisions. Firstly, we determined Sindhi stop-words by counting their term frequencies, and secondly, by analyzing their grammatical status with the help of Sindhi linguistic expert because all the frequent words are not stop-words (see Figure \ref{fig:freq_words}). After determining the importance of such words with human judgment, we placed them in the list of stop-words. The total number of detected stop-words is 340 in our developed corpus. The partial list of most frequent Sindhi stop-words is depicted in Table \ref{tab:stopwords} along with their frequency. The filtration of stop-words is an essential preprocessing step for learning GloVe \cite{pennington2014glove} word embedding. However, the sub-sampling approach  \cite{bojanowski2017enriching,mikolov2018advances} is used to discard such most frequent words in CBoW and SG models.
\begin{table}
	\centering
	\includegraphics[width=0.48\textwidth]{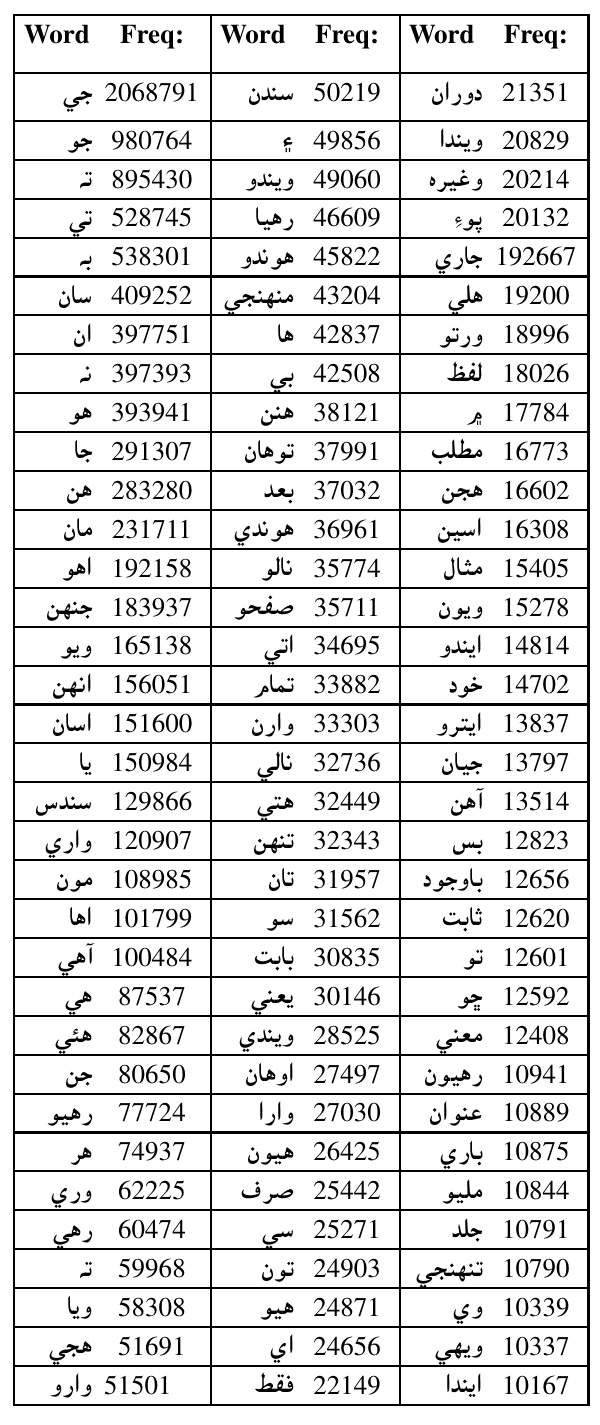}
		\caption{Partial list of  Sindhi stop-words in the corpus from most to less frequent. Freq: denotes the frequency of each word}
	\label{tab:stopwords}
\end{table}

\section{Word embedding models}
\label{sec:embedding_models}
The embedding models can be broadly categorized into predictive and count-based methods, being generated by employing co-occurrence statistics, NN based, and probabilistic algorithms. The GloVe \cite{pennington2014glove} treats each word as a single entity in the corpus and generates a vector of each word. However, CBoW and SG \cite{mikolov2013efficient,mikolov2013distributed}, later extended \cite{bojanowski2017enriching,mikolov2018advances}, rely on simple two-layered NN architecture which uses linear activation function in hidden layer and softmax in the output layer. 
\subsection{GloVe} 
\label{sec:glove}
The GloVe is a log-bilinear regression \cite{pennington2014glove} model, which combines two local context window and global matrix factorization methods. It weights the contexts using the harmonic function for training word embeddings from given input vocabulary in an unsupervised way. 

\subsection{Continuous bag-of-words} 
\label{subsec:CBoW}
The standard CBoW is the inverse of SG \cite{mikolov2013efficient} model, which predicts input word on behalf of the context. The length of an input to the CBoW model depends on the setting of context window size (\textit{ws}), which determines the distance to the left and right of the target word. Hence the context is a window that contains neighboring words. The objective of the CBoW is to maximize the probability of given neighboring words.

\subsection{Skip-gram} \label{sec:skipgram}
The SG model predicts surrounding words by giving input word \cite{mikolov2013distributed} with the training objective of learning good word embeddings that efficiently predict the neighboring words. The goal of skip-gram is to maximize average log-probability of words across the entire training corpus.

\section{Evaluation methods}
\label{sec:evaluation_matrix}
We employ intrinsic and extrinsic evaluation approaches for the performance analysis of proposed word embeddings. The intrinsic evaluation is based on semantic, syntactic similarity \cite{schnabel2015evaluation} in word embeddings. The word similarity approach states \cite{levy2015improving} that the words are similar if they appear in a similar context. The intrinsic evaluation is based on the nearest neighboring words, word pair relationship, and Sindhi WordSim-347.

For the extrinsic evaluation, we use the word embedding clusters
as features for two NLP tasks of part-of-speech tagging and named entity recognition. SiPOS~\cite{ali2021sipos} and SiNER~\cite{ali2020siner} datasets are used for experiments following the suggested standard split, respectively.   In the extrinsic evaluation of proposed word embeddings, we use them as input features to a downstream task of POS tagging and NER, measure the performance specific to that task. We exploit recently proposed neural model~\cite{ali2021sipos} based on bidirectional long short-term memory (BiLSTM) network, self-attention (SA), and sequential conditional random field (CRF). 
\section{Experiments and Results}
\label{sec:exp_set}
The embedding models are trained on the corpus after employing preprocessing pipeline to the crawled data. In the training phase, the optimal hyper-parameters \cite{schnabel2015evaluation} are more important than designing a novel algorithm. We optimize the dictionary and algorithm-based parameters of CBoW, SG, and GloVe algorithms.  All the experiments are conducted on the GTX 1080-TITAN GPU machine.
\subsection{Parameters}
\label{Hyperpara_optimization}
The state-of-the-art SG, CBoW \cite{mikolov2013efficient,mikolov2013distributed,bojanowski2017enriching,mikolov2018advances} and Glove~\cite{pennington2014glove} word embedding algorithms are exploited for the development of Sindhi word embeddings by parameter tuning. These parameters can be categorized into a dictionary and algorithm-based. Therefore, more robust embeddings became possible to train with hyper-parameters optimization.
\begin{table}[!b]
	\centering
	\begin{tabular}{|l|c|c|}
		\hline
		\textbf{Parameter}  & \textbf{CBoW}, \textbf{SG}   &  \textbf{GloVe} \\
		\hline
		Epoch   &   100     &  100 \\
		\hline
		Learning rate      &  0.25    &    0.25 \\	\hline
		Embedding dimension     &   300    &     300 \\ 	\hline
		minn char & 02    &    -- \\ 	\hline
		maxn char & 07   &     -- \\ 	\hline
		Window size      &   7      &    7 \\ 	\hline
		Negative sampling      &   20  &     -- \\ 	\hline
		minw    &   4   &     4 \\		\hline
	\end{tabular}  
	\caption{Parameters for CBoW, SG, and GloVe models}
	\label{tab:emb_parameters}
\end{table}
We tuned and evaluated the hyper-parameters of CBoW, SG, and GloVe algorithms individually, which are depicted in Table \ref{tab:emb_parameters}. The selection of minimum (\textit{minn}) and the maximum (\textit{maxn}) length of character $n-grams$ is an important parameter for learning character-level representations of words in CBoW and SG~\cite{bojanowski2017enriching}. Therefore, the n-grams from $3-9$ were tested to analyze the impact on the performance of embeddings. We optimized the length of character n-grams from $minn=2$ and $maxn=7$ by keeping in view the length of letter n-grams in words, depicted in Table \ref{tab:letter_ngrams}.  We evaluated the range of minimum word counts from 1 to 8. It is analyzed that the size of input vocabulary is decreasing at a large scale by ignoring more words. Similarly, the vocabulary size was increasing by considering rare words. Therefore, ignoring words with a frequency of $<5$ in CBoW, SG, and GloVe consistently yield better results. We use hierarchical softmax for CBoW, negative sampling for SG and default loss function for GloVe~\cite{pennington2014glove}. The recommended verbosity level, number of buckets, threads,  sampling threshold, are used for training CBoW, SG \cite{mikolov2018advances}, and GloVe \cite{pennington2014glove}.

\subsection{Intrinsic Evaluation}
The cosine similarity matrix~\cite{levy2015improving} is a common and primary method to measure the distance between a set of words and nearest neighbors. The high cosine similarity score denotes the closer words in the embedding matrix, while the less cosine similarity score means the higher distance between word pairs. The intrinsic evaluation methods of querying nearest neighboring words, word pair relationship, and WordSim-347. 
\begin{table*}
	\centering
	\includegraphics[width=16cm]{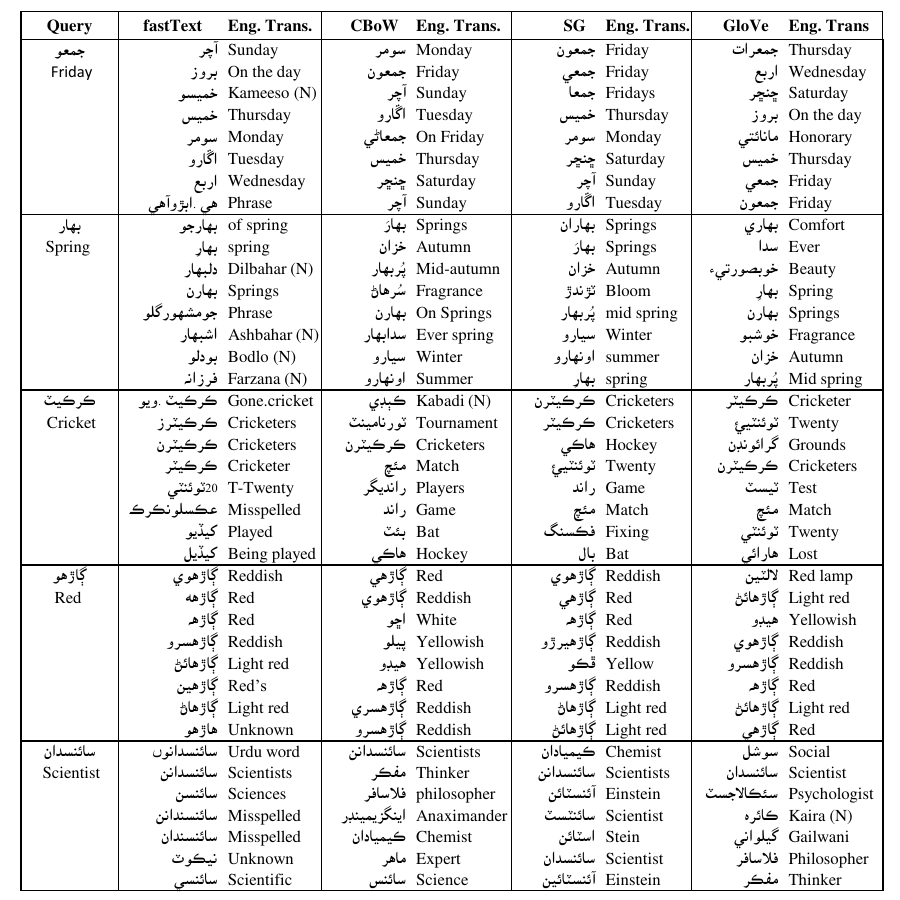}
	\caption{Eight nearest neighboring words of each query word. Eng. Trans. denotes the English translation of each word}
	\label{tab:nearest_neighnors}
\end{table*}
\label{subsec:word_similarity_comparision}
\subsubsection{Nearest neighboring words}
\label{sub:nearset_neighbors}
Each word contains the most similar top eight nearest neighboring words in Table \ref{tab:nearest_neighnors} determined by the highest cosine score.  We present English translation of both query and retrieved words also discuss their English meaning for ease of relevance judgment. To take a closer look at the semantic and syntactic relationship captured in the proposed word embeddings, Table \ref{tab:nearest_neighnors} shows the top eight nearest neighboring words of five different query words {\textbf{Friday (\includegraphics[width=0.59cm]{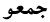}), Spring (\includegraphics[width=0.49cm]{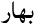}), Cricket (\includegraphics[width=1.1cm]{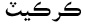}), Red (\includegraphics[width=0.68cm]{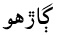}), Scientist (\includegraphics[width=1.1cm]{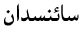})}} taken from the vocabulary. As the first query word \textit{Friday} returns the names of days \textit{Saturday, Sunday, Monday, Tuesday, Wednesday, Thursday} in an unordered sequence. The SdfastText returns five names of days \textit{Sunday, Thursday, Monday, Tuesday} and \textit{Wednesday}, respectively. The GloVe model also returns five names of days. However, CBoW and SG gave six names of days except \textit{Wednesday} along with different writing forms of query word being written in Sindhi language. The CBoW and SG returned more relevant words as compare to SdfastText and GloVe. Moreover,  the CBoW returned \textit{Add} and GloVe returns \textit{Honorary} words which are not similar to the query word. But SdfastText gave two irrelevant words of \textit{Kameeso (N)} which is a name (N) of a person in Sindhi, and \textit{Phrase} is a combination of three Sindhi words. Similarly, second query word \textit{Spring} retrieved accurately as names of seasons by CBoW, SG, and GloVe, respectively. However, SdfastText returned four irrelevant words of \textit{Dilbahar (N), Phrase, Ashbahar (N)} and \textit{Farzana (N)} out of eight nearest neighbors. The third query word \textit{Cricket} is the name of a popular game. The first retrieved word \textit{Kabadi (N)}, by CBoW is the name of a popular game being played in Asia. Including \textit{Kabadi (N)} all the returned words by CBoW, SG and GloVe are related to query word {\textit{Cricket}}. However, two words are combined with a punctuation mark (.) in the first retrieved word by SdfastText, which shows the tokenization error in preprocessing step. The sixth retrieved word \textit{Misspelled} is a combination of three words, not related to query word. Furthermore, the fourth query word \textit{Red} gave the names of closely related colors and different forms of query word being written in Sindhi language. The last returned word \textit{Unknown} by SdfastText is irrelevant, not found in Sindhi dictionary for translation. The query word {\textit{Scientist}} also contains semantically related words returned by CBoW, SG, and GloVe, but the first word given by SdfasText belongs to the Urdu language. Another \textit{unknown} word returned by SdfastText does not have any meaning in Sindhi dictionary. It seems that the vocabulary of SdfastText also contains words of other languages. More interesting observations in the presented results are the diacritized words retrieved by our proposed CBoW, SG, and GloVe word embedding models. Hence, the overall performance of the proposed SG, CBoW, and GloVe demonstrate high semantic relatedness in retrieving the top eight nearest neighbor words as compare to SdfastText.

\subsection{Word pair relationship}
\label{subsec:word_pair_relationship} 
The average similarity score between countries and their capitals is depicted in Table \ref{tab:country_capital} along with English translation of each word pair. The SG model yields the best average score of {\textbf{0.675}} followed by CBoW with a 0.634 similarity score. The GloVe also yields better semantic relatedness of 0.594, and the SdfastText yields an average score of 0.391, respectively. The first query word {\textbf{China-Beijing}} is not available in the vocabulary of SdfastText. It shows that along with performance, the vocabulary in SdfastText is also limited as compared to our proposed word embeddings.  However, the similarity score between {\textit{Afghanistan-Kabul}} is lower in our proposed CBoW, SG, and GloVe models, because the word {\textit{Kabul}} is the name of the capital of \textit{Afghanistan} as well as it frequently appears as an adjective in Sindhi text which means \textit{able}. 

%
\begin{table*}
	\centering
		\includegraphics[width=13cm]{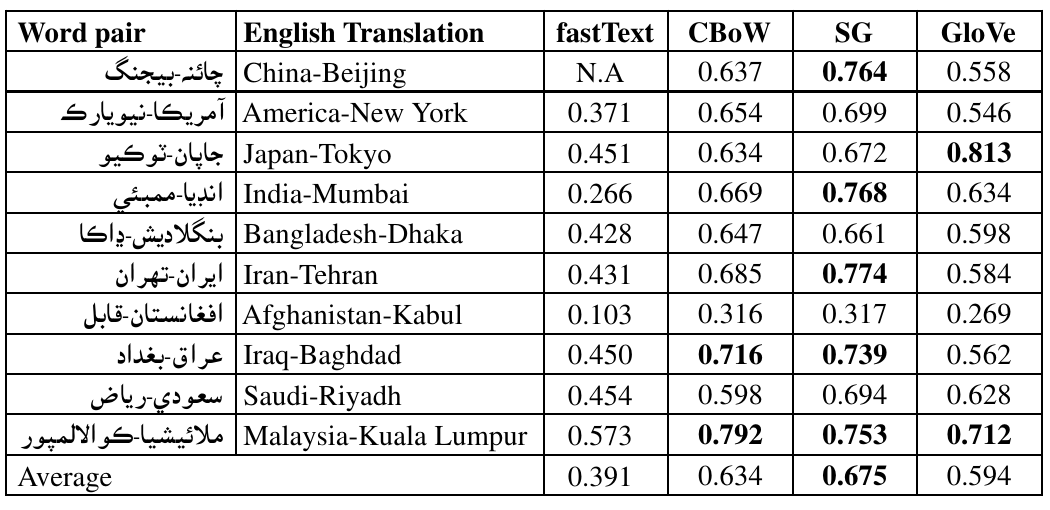}
		\caption{Cosine similarity score (higher is better) between country and capital. The bold results highlight the best scores between country and capital words}
	\label{tab:country_capital}
\end{table*}
\subsection{WordSim-347 }
\label{sec:comparision_wordsim-353}
We evaluate and compare the performance of our proposed word embeddings using the WordSim-353 dataset by translating English word pairs to subsequent Sindhi words. Due to vocabulary differences between English and Sindhi, we were unable to find the authentic meaning of six terms, so we left these terms untranslated. So our final Sindhi WordSim consists of 347 word pairs. Table \ref{tab:semantic_similarity} shows the Spearman correlation results on different dimensional embeddings on the translated Sindhi WordSim. The Table \ref{tab:semantic_similarity} shows complete results with the different \textit{ws} for CBoW, SG and GloVe. The window size of 7 subsequently yield better performance. The SG model outperforms CBoW and GloVe in semantic and syntactic similarity by achieving the accuracy of \textbf{0.651}. In comparison with English \cite{mikolov2013efficient} achieved the average semantic and syntactic similarity of 0.637, 0.656 with CBoW and SG, respectively. Moreover, CBoW, SG, and GloVe models also surpass the recently proposed SdfasText \cite{grave2018learning}. Therefore, despite the challenges in translation from English to Sindhi, our proposed word embeddings have efficiently captured the semantic and syntactic relationship.
\begin{table}
	\centering
		\begin{tabular}{|lcc|}
		\hline
		\textbf{Embedding Model} & \textbf{ws} & \textbf{Accuracy} \\ \hline
		\multirow{4}{*}{CBoW} & 3  & 0.596 \\
		& 5  & 0.618 \\
		& 7  & \textbf{0.621} \\ \hline
		\multirow{3}{*}{Skip gram} & 3 & \textbf{0.625} \\
		& 5 &  \textbf{0.649} \\
		& 7  & \textbf{0.651} \\ \hline
		\multirow{3}{*}{GloVe} & 3 & 0.593 \\
		& 5 &  0.614 \\
		& 7  & 0.618 \\ \hline
		SdfastText &  &  0.374 \\ \hline
	\end{tabular}
	\caption{Comparison of semantic and syntactic accuracy of proposed word embeddings using WordSim-353 dataset on $300-D$ embedding, on choosing various window sizes}
	\label{tab:semantic_similarity}
\end{table}
\subsection{Embedding visualization}
\label{sec:visualization}
The purpose of embeddings visualization is to keep similar words close together in 2-dimensional $x,y$ coordinate pairs while maximizing the distance between dissimilar words. The t-SNE has a tunable perplexity (PPL) parameter to balance the data points at both the local and global levels.
\begin{figure}
	\centering
	\includegraphics[width=8cm]{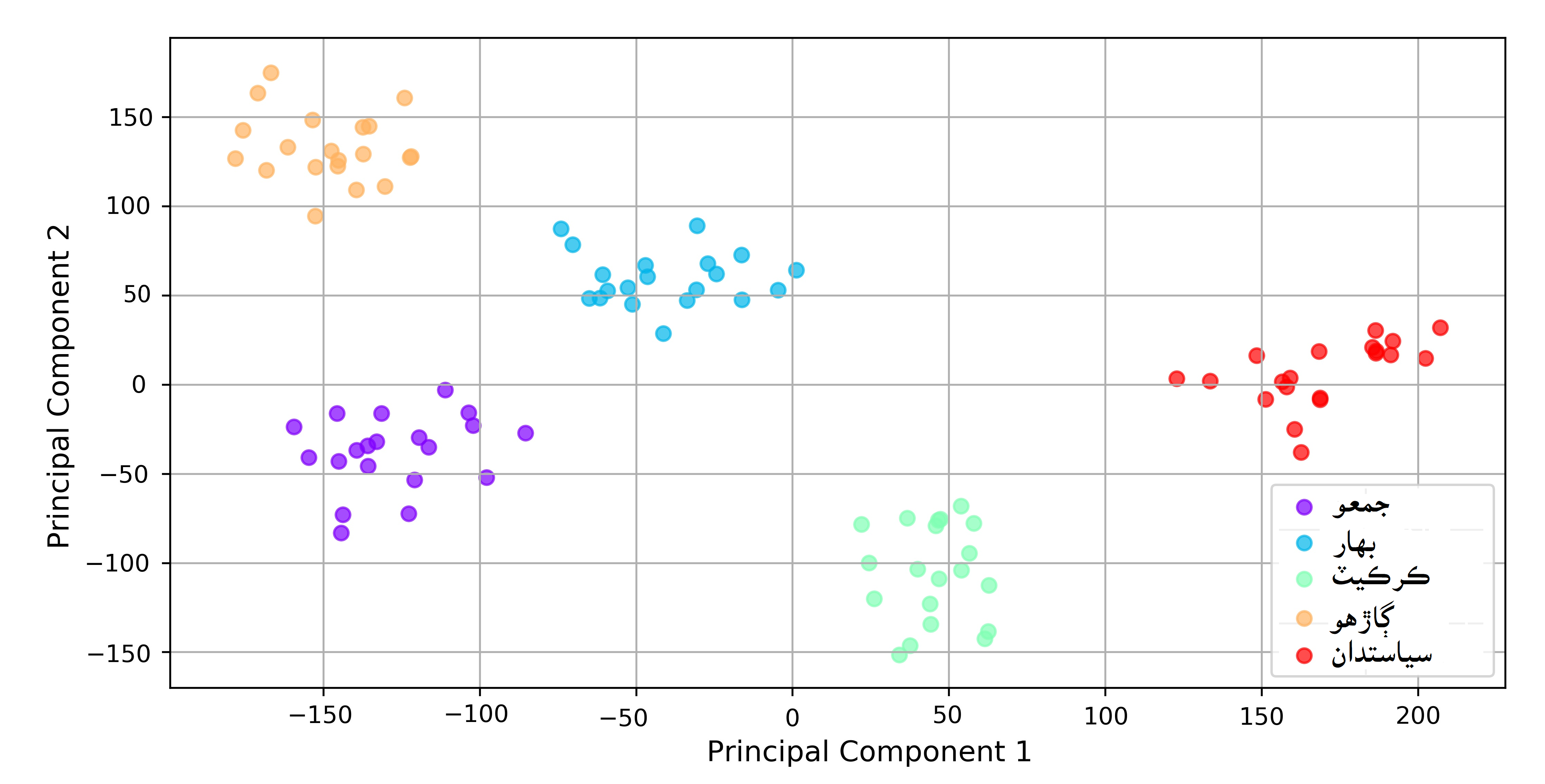}
	\caption{Visualization of Sindhi CBoW word embeddings}
	\label{fig:viz_cbow}
\end{figure}
\begin{figure}
	\centering
	\includegraphics[width=8cm]{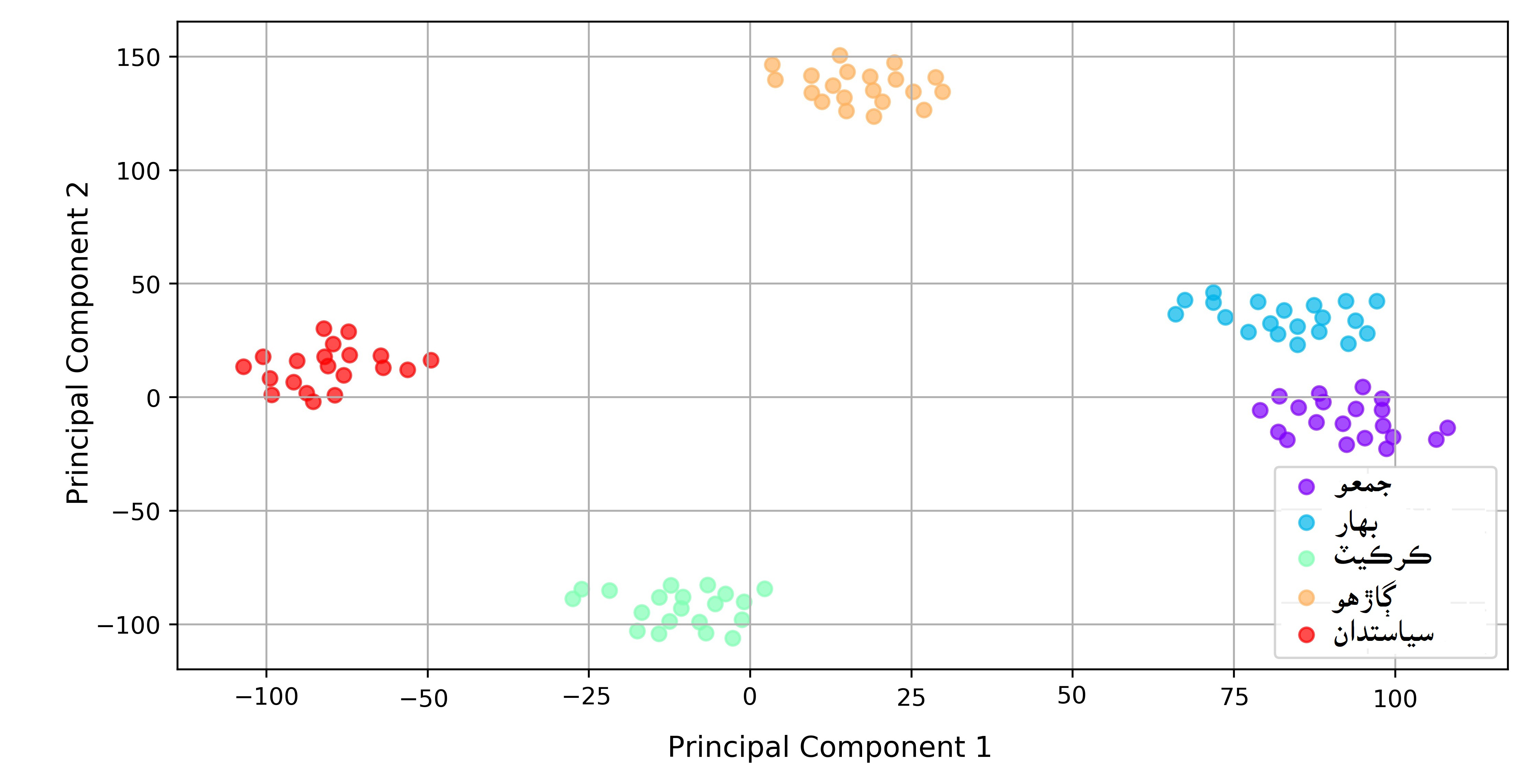}
	\caption{Visualization of the Sindhi SG word embeddings}
	\label{fig:viz_SG}
\end{figure}
\begin{figure}
	\centering
	\includegraphics[width=8cm]{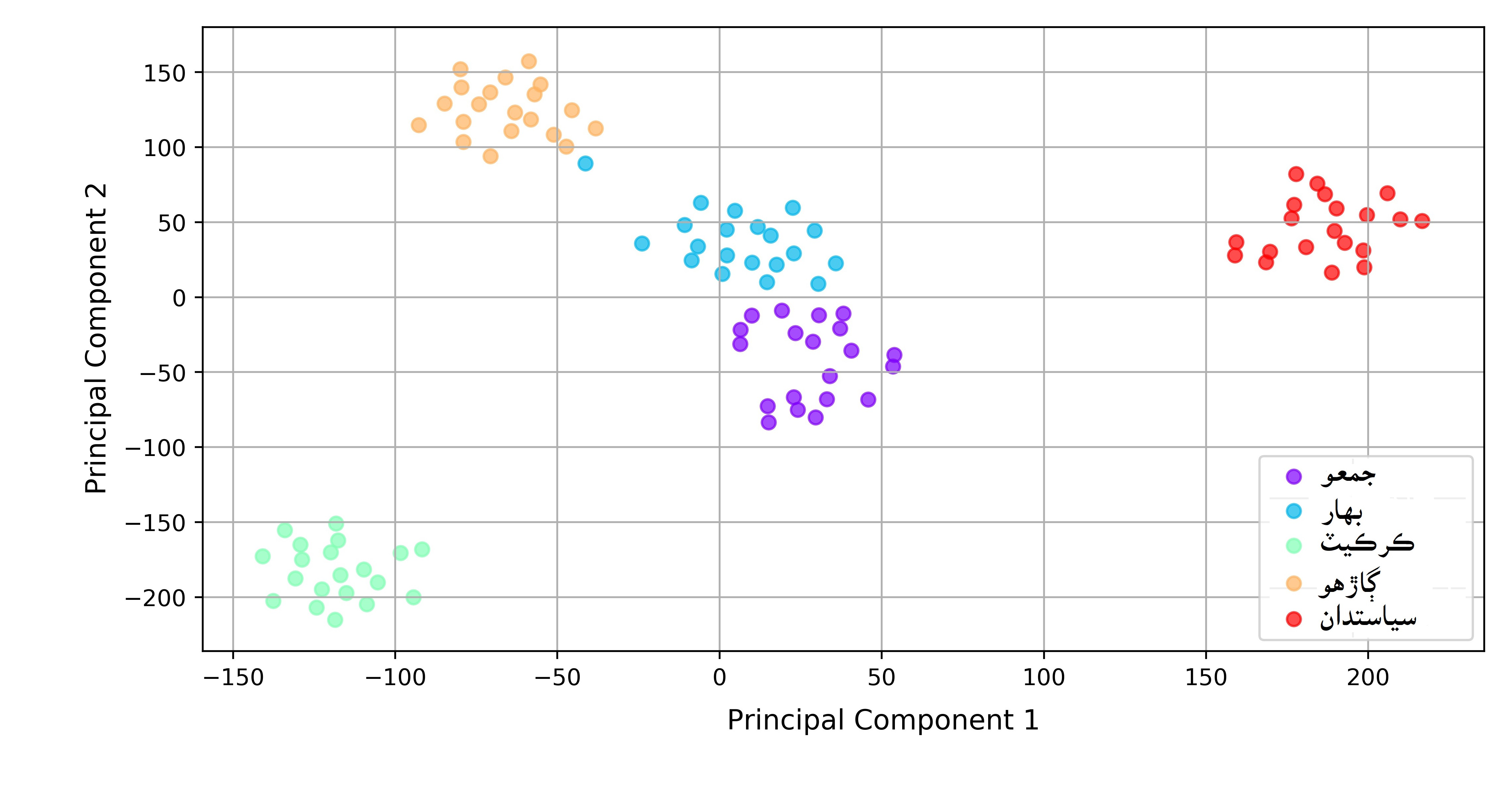}
	\caption{Visualization of the Sindhi GloVe word embeddings}
	\label{fig:viz_glove}
\end{figure}
\begin{figure}
	\centering
	\includegraphics[width=8cm]{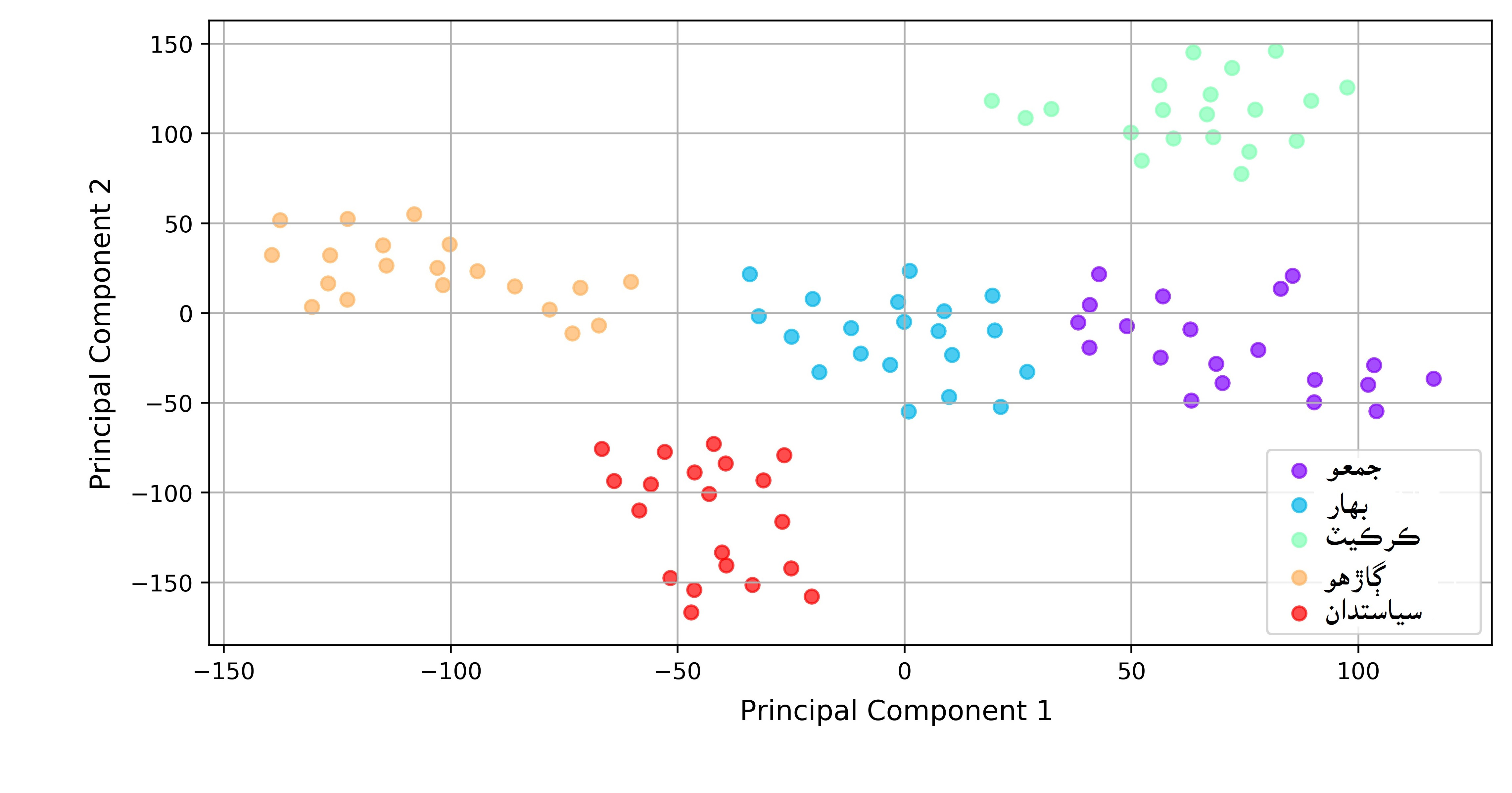}
	\caption{Visualization of the SdfastText word embeddings}
	\label{fig:viz_fasttext}
\end{figure}
We visualize the 300-\textit{D} embeddings using $PPL=20$ on 5000-iterations. We use the same query words (see Table \ref{tab:nearest_neighnors}) by retrieving the top 20 nearest neighboring word clusters for a better understanding of the distance between similar words. Every query word has a distinct color for the clear visualization of a similar group of words. The closer word clusters show the high similarity between the query and retrieved word clusters. The word clusters in SG (see Fig. \ref{fig:viz_SG}) are closer to their group of semantically related words. Secondly, the CBoW model depicted in Fig. \ref{fig:viz_cbow} and GloVe (Fig. \ref{fig:viz_glove}) also show the better cluster formation of words than SdfastText (Fig. \ref{fig:viz_fasttext}), respectively.

\subsection{Extrinsic Evaluation}
AdaGrad is used for training statistical models. As recommended by Levy, Goldberg, and
Dagan (2015), additional experiments are conducted by concatenating the word and contextual vectors (w+c).
For the NER experiments, the highest F1-score of 86.19 is
achieved by the skip-gram with negative sampling embeddings (SGNS) using the agglomerative clustering. On the
other hand, the highest accuracy of 97.51 is achieved by
Brown clustering (using raw text instead of embeddings).
These results outperform the previous work (Pradhan et al.
2013), showing the absolute improvements of 3.77
0.42
All of the above experiments are using the maximum cluster size of 1,500. We also tested on the max cluster size of 15,000, which showed very similar results. This implies that
the increase of cluster size does not improve the quality of the
clusters, at least for these two tasks. For the NER task, SGNS
and Brown give constant additive increase in performance
regardless of the size of the training data

\begin{table}[h]
	\centering
		\begin{tabular}{|l|c|}
		\hline
		\textbf{Embedding Model} & \textbf{Accuracy} \\ \hline
	    CBoW & 95.17   \\ \hline
		SG & 95.32   \\ \hline
		GloVe & 94.52 \\  \hline
		SdfastText & 92.76  \\ \hline
	\end{tabular}
	\caption{Extrinsic evaluation of various embeddings on Sindhi POS tagging task using SiPOS dataset}
	\label{tab:sdner}
\end{table}

\begin{table}[h]
	\centering
		\begin{tabular}{|l|c|c|c|}
		\hline
		\textbf{Embedding Model} & \textbf{P\%} & \textbf{R\%} & \textbf{F1\%} \\ \hline
	    CBoW & 90.18 & 90.39 & 89.9 \\ \hline
		SG & 90.27 & 90.64 & 90.11 \\ \hline
		GloVe & 88.37 & 89.25 & 88.79 \\ \hline
		fastText & 84.  & 84. & 84. \\ \hline
	\end{tabular}
	\caption{Extrinsic evaluation of various embeddings on Sindhi NER task using SiNER dataset}
	\label{tab:sdner}
\end{table}

\begin{table}[h]
	\centering
		\begin{tabular}{|l|c|c|c|}
		\hline
		\textbf{Embedding Model} & \textbf{P\%} & \textbf{R\%} & \textbf{F1\%} \\ \hline
	    CBoW & 0.18 & 00.39 & 00.9 \\ \hline
		SG & 0.27 & 00.64 & 00.11 \\ \hline
		GloVe & 0.37 & 00.25 & 00.79 \\ \hline
		fastText & .00  & 00. & 00. \\ \hline
	\end{tabular}
	\caption{Extrinsic evaluation of various pretrained word embeddings using BiLSTM-SA model on Sindhi NER task using SiNER dataset}
	\label{tab:sdner}
\end{table}

\section{Conclusion and Future Work}
\label{sec:conclusion}
In this paper, we propose word embedding based large corpus of more than 61 million tokens, list of stop-words, and Sindhi word embeddings using state-of-the-art CBoW, SG, and GloVe algorithms. The proposed word embeddings are evaluated using the popular intrinsic and extrinsic evaluation approaches. We translate English WordSim-353 to Sindhi by using the English-Sindhi bilingual dictionary for intrinsic evaluation. The empirical results demonstrate that SG yields the best performance than CBoW and GloVe models by retrieving authentic nearest neighboring words, a high average cosine similarity score of 0.660 between different word pairs, 0.675 between country and capital words, and 0.651 accuracy on WordSim-347. 

In the future, we will utilize the corpus for annotation projects such as POS tagging named entity recognition and sentiment analysis.  Moreover, the generated word embeddings can be utilized for the automatic construction of Sindhi WordNet. Furthermore, we will also utilize the corpus using Bi-directional Encoder Representation Transformer BERT, bi-directional language model Elmo and Generative Pretrained Transformer GPT for learning deep contextualized representations. 
%



\section{Bibliographical References}\label{reference}

\bibliographystyle{lrec2022-bib}
\bibliography{lrec2022}

\begin{thebibliography}{}

\bibitem[\protect\citename{Ali and Wagan}2017]{ali2017sentiment}
Ali, M. and Wagan, A.~I.
\newblock (2017).
\newblock Sentiment summerization and analysis of {Sindhi} text.
\newblock {\em Int. J. Adv. Comput. Sci. Appl}, 8(10):296--300.

\bibitem[\protect\citename{Ali \bgroup et al.\egroup }2015]{ali2015towards}
Ali, W., Kehar, A., and Shaikh, H.
\newblock (2015).
\newblock Towards {Sindhi} named entity recognition: Challenges and
  opportunities.
\newblock In {\em 1st National Conference on Trends and Innovations in
  Information Technology}.

\bibitem[\protect\citename{Ali \bgroup et al.\egroup }2020]{ali2020siner}
Ali, W., Lu, J., and Xu, Z.
\newblock (2020).
\newblock Siner: A large dataset for {Sindhi} named entity recognition.
\newblock In {\em Proceedings of The 12th Language Resources and Evaluation
  Conference}, pages 2953--2961.

\bibitem[\protect\citename{Ali \bgroup et al.\egroup }2021a]{ali2021creating}
Ali, W., Ali, N., Dai, Y., Kumar, J., Tumrani, S., and Xu, Z.
\newblock (2021a).
\newblock Creating and evaluating resources for sentiment analysis in the
  low-resource language: Sindhi.
\newblock In {\em Proceedings of the Eleventh Workshop on Computational
  Approaches to Subjectivity, Sentiment and Social Media Analysis}, pages
  188--194.

\bibitem[\protect\citename{Ali \bgroup et al.\egroup }2021b]{ali2021sipos}
Ali, W., Xu, Z., and Kumar, J.
\newblock (2021b).
\newblock Sipos: A benchmark dataset for sindhi part-of-speech tagging.
\newblock In {\em Proceedings of the Student Research Workshop Associated with
  RANLP 2021}, pages 22--30.

\bibitem[\protect\citename{Bhatti \bgroup et al.\egroup }2014]{bhatti2014word}
Bhatti, Z., Ismaili, I.~A., Soomro, W.~J., and Hakro, D.~N.
\newblock (2014).
\newblock Word segmentation model for {Sindhi} text.
\newblock {\em American Journal of Computing Research Repository}, 2(1):1--7.

\bibitem[\protect\citename{Bird \bgroup et al.\egroup }2009]{nltk}
Bird, S., Klein, E., and Loper, E.
\newblock (2009).
\newblock {\em Natural language processing with Python: analyzing text with the
  natural language toolkit}.
\newblock " O'Reilly Media, Inc.".

\bibitem[\protect\citename{Bojanowski \bgroup et al.\egroup
  }2017]{bojanowski2017enriching}
Bojanowski, P., Grave, E., Joulin, A., and Mikolov, T.
\newblock (2017).
\newblock Enriching word vectors with subword information.
\newblock {\em Transactions of the Association for Computational Linguistics},
  5:135--146.

\bibitem[\protect\citename{Che \bgroup et al.\egroup }2010]{che2010ltp}
Che, W., Li, Z., and Liu, T.
\newblock (2010).
\newblock Ltp: A chinese language technology platform.
\newblock In {\em Proceedings of the 23rd International Conference on
  Computational Linguistics: Demonstrations}, pages 13--16. Association for
  Computational Linguistics.

\bibitem[\protect\citename{Cole}2006]{cole2006sindhi}
Cole, J.
\newblock (2006).
\newblock Sindhi. encyclopedia of language \& linguistics volume8.

\bibitem[\protect\citename{Dootio and Wagan}2017]{dootio2017automatic}
Dootio, M.~A. and Wagan, A.~I.
\newblock (2017).
\newblock Automatic stemming and lemmatization process for {Sindhi} text.
\newblock {\em J. Soc. Sci. Interdiscip. Res.(JSSIR), NED Univ. Eng. Technol.
  Karachi Sindh Pakistan}, 6(2):19--28.

\bibitem[\protect\citename{Dootio and Wagan}2019a]{dootio2019development}
Dootio, M.~A. and Wagan, A.~I.
\newblock (2019a).
\newblock Development of {Sindhi} text corpus.
\newblock {\em Journal of King Saud University-Computer and Information
  Sciences}.

\bibitem[\protect\citename{Dootio and Wagan}2019b]{dootio2019syntactic}
Dootio, M.~A. and Wagan, A.~I.
\newblock (2019b).
\newblock Syntactic parsing and supervised analysis of {Sindhi} text.
\newblock {\em Journal of King Saud University-Computer and Information
  Sciences}, 31(1):105--112.

\bibitem[\protect\citename{Grave \bgroup et al.\egroup
  }2018]{grave2018learning}
Grave, E., Bojanowski, P., Gupta, P., Joulin, A., and Mikolov, T.
\newblock (2018).
\newblock Learning word vectors for 157 languages.
\newblock In {\em Proceedings of the Eleventh International Conference on
  Language Resources and Evaluation (LREC-2018)}.

\bibitem[\protect\citename{Honnibal and Montani}2017]{spacy}
Honnibal, M. and Montani, I.
\newblock (2017).
\newblock Natural language understanding with bloom embeddings, convolutional
  neural networks and incremental parsing.
\newblock {\em Unpublished software application. https://spacy. io}.

\bibitem[\protect\citename{Khodak \bgroup et al.\egroup
  }2017]{khodak2017automated}
Khodak, M., Risteski, A., Fellbaum, C., and Arora, S.
\newblock (2017).
\newblock Automated {WordNet} construction using word embeddings.
\newblock In {\em Proceedings of the 1st Workshop on Sense, Concept and Entity
  Representations and their Applications}, pages 12--23.

\bibitem[\protect\citename{Khoso \bgroup et al.\egroup }2019]{khoso2019build}
Khoso, F.~H., Memon, M.~A., Nawaz, H., and Musavi, S. H.~A.
\newblock (2019).
\newblock To build corpus of {Sindhi}.

\bibitem[\protect\citename{Levy \bgroup et al.\egroup }2015]{levy2015improving}
Levy, O., Goldberg, Y., and Dagan, I.
\newblock (2015).
\newblock Improving distributional similarity with lessons learned from word
  embeddings.
\newblock {\em Transactions of the Association for Computational Linguistics},
  3:211--225.

\bibitem[\protect\citename{Manning \bgroup et al.\egroup
  }2014]{manning2014stanford}
Manning, C., Surdeanu, M., Bauer, J., Finkel, J., Bethard, S., and McClosky, D.
\newblock (2014).
\newblock The {Stanford} {CoreNLP} natural language processing toolkit.
\newblock In {\em Proceedings of 52nd annual meeting of the association for
  computational linguistics: system demonstrations}, pages 55--60.

\bibitem[\protect\citename{Mikolov \bgroup et al.\egroup
  }2013a]{mikolov2013efficient}
Mikolov, T., Chen, K., Corrado, G., and Dean, J.
\newblock (2013a).
\newblock Efficient estimation of word representations in vector space.
\newblock {\em arXiv preprint arXiv:1301.3781}.

\bibitem[\protect\citename{Mikolov \bgroup et al.\egroup
  }2013b]{mikolov2013distributed}
Mikolov, T., Sutskever, I., Chen, K., Corrado, G.~S., and Dean, J.
\newblock (2013b).
\newblock Distributed representations of words and phrases and their
  compositionality.
\newblock In {\em Advances in neural information processing systems}, pages
  3111--3119.

\bibitem[\protect\citename{Mikolov \bgroup et al.\egroup
  }2018]{mikolov2018advances}
Mikolov, T., Grave, E., Bojanowski, P., Puhrsch, C., and Joulin, A.
\newblock (2018).
\newblock Advances in pre-training distributed word representations.
\newblock In {\em Proceedings of the Eleventh International Conference on
  Language Resources and Evaluation (LREC-2018)}.

\bibitem[\protect\citename{Motlani}2016]{motlani2016developing}
Motlani, R.
\newblock (2016).
\newblock Developing language technology tools and resources for a
  resource-poor language: {Sindhi}.
\newblock In {\em Proceedings of the NAACL Student Research Workshop}, pages
  51--58.

\bibitem[\protect\citename{Nathani \bgroup et al.\egroup
  }2019]{nathani2019design}
Nathani, B., Joshi, N., and Purohit, G.
\newblock (2019).
\newblock Design and development of lemmatizer for {Sindhi} language in
  {Devanagri} script.
\newblock {\em Journal of Statistics and Management Systems}, 22(4):635--641.

\bibitem[\protect\citename{Nayak \bgroup et al.\egroup
  }2016]{nayak2016evaluating}
Nayak, N., Angeli, G., and Manning, C.~D.
\newblock (2016).
\newblock Evaluating word embeddings using a representative suite of practical
  tasks.
\newblock In {\em Proceedings of the 1st Workshop on Evaluating Vector-Space
  Representations for NLP}, pages 19--23.

\bibitem[\protect\citename{Otter \bgroup et al.\egroup }2020]{otter2020survey}
Otter, D.~W., Medina, J.~R., and Kalita, J.~K.
\newblock (2020).
\newblock A survey of the usages of deep learning for natural language
  processing.
\newblock {\em IEEE Transactions on Neural Networks and Learning Systems}.

\bibitem[\protect\citename{Padr{\'o} \bgroup et al.\egroup
  }2010]{padro2010freeling}
Padr{\'o}, L., Collado, M., Reese, S., Lloberes, M., and Castell{\'o}n, I.
\newblock (2010).
\newblock Freeling 2.1: Five years of open-source language processing tools.
\newblock In {\em 7th International Conference on Language Resources and
  Evaluation}.

\bibitem[\protect\citename{Pandey and Siddiqui}2009]{pandey2009evaluating}
Pandey, A.~K. and Siddiqui, T.~J.
\newblock (2009).
\newblock Evaluating effect of stemming and stop-word removal on {Hindi} text
  retrieval.
\newblock In {\em Proceedings of the First International Conference on
  Intelligent Human Computer Interaction}, pages 316--326. Springer.

\bibitem[\protect\citename{Pennington \bgroup et al.\egroup
  }2014]{pennington2014glove}
Pennington, J., Socher, R., and Manning, C.
\newblock (2014).
\newblock Glove: {Global} vectors for word representation.
\newblock In {\em Proceedings of the 2014 conference on empirical methods in
  natural language processing (EMNLP)}, pages 1532--1543.

\bibitem[\protect\citename{Pierrejean and Tanguy}2018]{pierrejean2018towards}
Pierrejean, B. and Tanguy, L.
\newblock (2018).
\newblock Towards qualitative word embeddings evaluation: Measuring neighbors
  variation.
\newblock In {\em Proceedings of the 2018 Conference of the North American
  Chapter of the Association for Computational Linguistics: Student Research
  Workshop}, pages 32--39.

\bibitem[\protect\citename{Popel and
  {\v{Z}}abokrtsk{\`y}}2010]{popel2010tectomt}
Popel, M. and {\v{Z}}abokrtsk{\`y}, Z.
\newblock (2010).
\newblock Tectomt: modular nlp framework.
\newblock In {\em International Conference on Natural Language Processing},
  pages 293--304. Springer.

\bibitem[\protect\citename{Rahman}2010]{rahman2010towards}
Rahman, M.~U.
\newblock (2010).
\newblock Towards {Sindhi} corpus construction.
\newblock In {\em Conference on Language and Technology, Lahore, Pakistan}.

\bibitem[\protect\citename{Sch{\"a}fer and Bildhauer}2013]{schafer2013web}
Sch{\"a}fer, R. and Bildhauer, F.
\newblock (2013).
\newblock Web corpus construction.
\newblock {\em Synthesis Lectures on Human Language Technologies}, 6(4):1--145.

\bibitem[\protect\citename{Schnabel \bgroup et al.\egroup
  }2015]{schnabel2015evaluation}
Schnabel, T., Labutov, I., Mimno, D., and Joachims, T.
\newblock (2015).
\newblock Evaluation methods for unsupervised word embeddings.
\newblock In {\em Proceedings of the 2015 Conference on Empirical Methods in
  Natural Language Processing}, pages 298--307.

\bibitem[\protect\citename{Shah \bgroup et al.\egroup }2018]{shah2018designing}
Shah, S. M.~A., Ismaili, I.~A., Bhatti, Z., and Waqas, A.
\newblock (2018).
\newblock Designing {XML} tag based {Sindhi} language corpus.
\newblock In {\em 2018 International Conference on Computing, Mathematics and
  Engineering Technologies (iCoMET)}, pages 1--5. IEEE.

\end{thebibliography}

\end{document}